\documentclass{article}

\usepackage[nonatbib,preprint]{neurips_2021}
\usepackage[numbers]{natbib}

\usepackage[utf8]{inputenc} 
\usepackage[T1]{fontenc}    
\usepackage[]{hyperref}       
\usepackage{url}            
\usepackage{booktabs}       
\usepackage{siunitx}        
\usepackage{amsfonts}       
\usepackage{nicefrac}       
\usepackage{microtype}      

\usepackage{makecell}
\usepackage{amsmath,amsfonts,amssymb,bbm}
\usepackage[pdftex]{graphicx}
\usepackage{wrapfig}
\usepackage{tikz}
\usepackage{rotating}
\usepackage{subcaption}

\usepackage{xr}

\usepackage{wasysym}
\usepackage{marvosym}

\usepackage{algorithm}
\usepackage{algorithmic}
\usepackage{setspace}

\usepackage{listings}
\usepackage{placeins}

\usepackage{xcolor}
\definecolor{codegreen}{rgb}{0,0.6,0}
\definecolor{codegray}{rgb}{0.5,0.5,0.5}
\definecolor{codepurple}{rgb}{0.58,0,0.82}
\definecolor{backcolour}{rgb}{0.95,0.95,0.92}

\definecolor{darkgreen}{rgb}{0,0.4,0}
\definecolor{cerise}{rgb}{0.871, 0.192, 0.388}
\definecolor{carmine}{rgb}{0.59, 0.0, 0.09}
\newtheorem{conjecture}{Conjecture}

\def\din{d_{\mathrm{in}}}
\def\dout{d_{\mathrm{out}}}
\def\DBN{\mathcal{D}_{\mathrm{BN}}}

\def\X{\mathcal{X}}

\def\Y{\mathcal{Y}}
\def\D{\mathcal{D}}

\def\R{\mathbb{R}}

\title{A Generalized Lottery Ticket Hypothesis}

\author{%
    \centerline{Ibrahim Alabdulmohsin$^{\star\dagger}$, Larisa Markeeva$^\ddagger$, Daniel Keysers$^\dagger$, Ilya Tolstikhin$^{\star\dagger}$
}
    \vspace{3mm}\\
    \centerline{$^\dagger$Google Research, Brain Team
    \hspace{.5cm}
    $^\ddagger$DeepMind}
    \vspace{3mm} \\
    \centerline{\texttt{{\{ibomohsin,lmarkeeva,keysers,tolstikhin\}}@google.com}}  \vspace{1.5mm}
}

\makeatletter
\newcommand*{\addFileDependency}[1]{
  \typeout{(#1)}
  \@addtofilelist{#1}
  \IfFileExists{#1}{}{\typeout{No file #1.}}
}
\makeatother

\begin{document}

\maketitle

\begin{abstract}
We introduce a generalization to the lottery ticket hypothesis in which the notion of ``sparsity'' is relaxed by choosing an arbitrary basis in the space of parameters.
We present evidence that the original results reported for the canonical basis continue to hold in this broader setting.
We describe how \emph{structured} pruning methods, including pruning units or factorizing fully-connected layers into products of low-rank matrices, can be cast as particular instances of this ``generalized'' lottery ticket hypothesis. The investigations reported here are preliminary and are provided to encourage further research along this direction.
\end{abstract}

\section{Introduction}
Overparameterization in deep neural networks (DNNs) simplifies training and improves generalization~\citep{kaplan2020scaling,du2018gradient,bartlett2021deep}. Nevertheless, overparameterized models may exhibit limitations in practice, such as on mobile devices, due to their large memory footprint and inference time~\citep{hoefler2021sparsity}. One remedy is pruning: \citet{hoefler2021sparsity} estimate that pruning can potentially improve the computational and storage efficiency of deep neural networks by up to two orders of magnitude. 

Driven by such goals, interest in model pruning (a.k.a.\ compression or sparsity) has grown. A fundamental premise behind pruning methods is that overparameterization leads to redundancies that can be compressed without impacting the overall performance. For example, \citet{denil2013predicting} show that 95\% of the model parameters in modern neural networks can be predicted from the remaining 5\%, while \citet{frankle2018lottery} show that over 95\% of the model parameters can be pruned without impacting the accuracy, a phenomenon that is referred to as the ``Lottery Ticket Hypothesis.'' Some of the observations often noted under the Lottery Ticket Hypothesis are summarized as follows: (1)~over 95\% compression can be achieved without impacting the test accuracy, (2)~the test accuracy improves initially with pruning before it begins to drop, and (3)~there exists a critical compression threshold after which the test accuracy decreases rapidly. These findings suggest that significant savings in memory footprint and/or inference time can be realized for deep neural networks.

However, the Lottery Ticket Hypothesis considers \textit{unstructured} sparsity. In practice, unstructured sparsity may not lead to significant speedups as accelerators are optimized for dense matrix operations~\citep{han2016eie}. Taking inference time as a metric, for example, \citet{wen2016learning} present experiments which show that the speedup in unstructured pruning is  limited even when the level of sparsity exceeds 95\%. 

In this paper, we present preliminary investigations into a ``Generalized Lottery Ticket Hypothesis'', which encompasses both unstructured and structured pruning under a common framework. Specifically, we 
notice that the notion of sparsity employed in the original works on the lottery ticket phenomenon (``many DNN weights are zero'') stems from the particular choice of the basis in the DNN parameter space (canonical basis).
We present evidence that the same phenomenon continues to hold for other, arbitrary choices of the basis (and their implied notions of the sparsity), resulting in the ``Generalized Lottery Ticket Hypothesis''.
We highlight two implications. First, one can achieve better compression using a more appropriate basis, such as the discrete cosine transform (DCT).
Second, one can achieve \textit{structured} pruning by carefully selecting the basis and using algorithms originally developed for unstructured sparsity, such as the iterative magnitude pruning (IMP) method.

\section{Generalized Lottery Ticket Hypothesis}
\paragraph{Notations.}If $J\subseteq\R^d$ is a subspace, we write $P_J$ for its projection matrix (i.e. the unique matrix satisfying $P_J=P_J^T$, $P_J^2=P_J$, and $P_Jx=x\leftrightarrow x\in J$).  Let $w_0\in\R^d$ be  an initialization of some deep neural network (DNN) architecture, which is trained using the stochastic gradient descent (SGD) method. We say that a model is \emph{trained over the subspace} $J$ if the initialization $w_0$ is projected onto $J$ and all the subsequent steps are confined within $J$. For example, pruning individual model parameters would correspond to training over a subspace $J$ spanned by a subset of the canonical basis $\{e_j\}_{j=1,\ldots,d}$, where $e_j$ is zero everywhere except at the $j$th entry.

\paragraph{Formal Definitions.}We begin by formally stating the ``Generalized Lottery Ticket Hypothesis.'' Consider an input space $\X$ (e.g.\ natural images)
and a finite output space $\Y=\{1,\dots,K\}$. 
Consider a deep neural network (DNN) with a fixed architecture.
We will denote a vector of the DNN's $d$~parameters by $w\in\R^d$. Such a
DNN architecture implements a function $y = f(x; w)$ that maps an input $x\in \X$ to an output $y \in \Y$. 

Often, the parameter vector $w$ is learned using a variant of the SGD method starting with some initial value $w_0\in\R^d$ and leading to a solution that is \emph{dense}; i.e. $||w||_0\approx d$. 
We refer to this training procedure as \emph{vanilla training}.
The Lottery Ticket Hypothesis \cite{frankle2018lottery} postulates that for any fixed initialization $w_0$, there exists 
a low-dimensional subspace $J$ spanned by $s\ll d$ elements of the canonical basis $\{e_i\}_{i=1,\ldots,d}$,
such that the following holds: training the DNN architecture over $J$ leads to a solution whose test accuracy is as good as the vanilla training. 

Instead of focusing on the canonical basis, we consider an arbitrary set of orthonormal vectors.
We refer 
to a set of orthonormal vectors $\D:=\{v_1,\dots, v_d\}\subset\R^d$ in the parameter space as a \emph{dictionary}.
For a dictionary $\D$ and a fixed vector $w\in\R^d$, we say that $w$ is $s$-\emph{sparse with respect to} $\D$ if
\[
w = \sum_{i=1}^d w_i v_i,\quad \text{with} \quad \|w\|_{\D,0} := \sum_{i=1}^d \mathbbm{1}\{ w_i \neq 0 \} \le s.
\]
In other words, $w$ is sparse w.r.t. $\D$ if and  only if a few basis vectors in $\D$ are sufficient to express $w$.

Different choices of $\D$ lead to different \emph{semantic meanings} of sparsity.
For the canonical basis, for example, sparsity implies ``many zero model parameters.''
As will be shown later, there exist choices of $\D$ where sparsity implies ``many coordinates of the input are completely ignored'' or ``many units in the given layer are removed,'' and so on.

With this, we are ready to state the Generalized Lottery Ticket Hypothesis (GLTH):
\begin{conjecture}[GLTH]
Fix any DNN architecture with $d$ trainable parameters.
For any dictionary $\D\subset\mathbb{R}^d$, there exists a subset $\D'\subsetneq\D$ with $|\D'|\ll d$ such that training the DNN model over the subspace $\textbf{Span}(\D')$ yields a solution whose test accuracy is, at least, as large as the test accuracy of vanilla training.
\end{conjecture}

Original works on the Lottery Ticket Hypothesis confirm (empirically) a special case of the GLTH above where $\D$ is the canonical basis of $\R^d$.
In that case, \citet{frankle2018lottery} propose the Iterative Magnitude Pruning (IMP) algorithm to find such sparse solutions, which is analogous to the classical orthogonal matching pursuit~(OMP) in signal processing~\citep{mallat1993matching}. We show, next, how IMP can be extended to the GLTH as well.

Assume we have a fixed DNN architecture $f(x; w)$ with $d$ trainable parameters and a dictionary  $\D=\{v_1,\dots, v_d\} \subset\R^d$.
The \emph{IMP algorithm} proceeds as follows:
\begin{enumerate}
    \item Sample a random initialization $w_0 \in \R^d$. Set $\D_1=\D$;
    \item For rounds $t = 1,\dots, T$:
    \begin{enumerate}
        \item {\em Rewind and Re-train}: Starting with $P_{\textbf{Span}(\D_{t})}w_0$, train the model parameters over the subspace $\textbf{Span}(\D_t)$ to convergence. Let $w_t$ be the new solution;
        \item {\em Sparsify}: Remove one (or more) elements from $\D_t$, resulting in a new smaller dictionary $\D_{t+1}\subsetneq \D_t$, where $\D_{t+1}$ is chosen to solve the following objective:
        \[
        \min_{\D_{t+1}\subsetneq \D_t:\; |\D_{t+1}|\le \tau |\D_{t}|} ||w_t-P_{\textbf{Span}(\D_{t+1})}w_t ||_{2}^2.
        \]
        The hyperparameter $\tau\in(0, 1)$ controls the fraction of elements in the remaining dictionary to be removed.
    \end{enumerate}
\end{enumerate}
For a fixed dictionary $\D$ any vector of model parameters $w\in\R^d$ can be re-parametrized using the corresponding orthonormal basis $w = a_1 v_1 + \dots + a_d v_d$.
In practice on step 2.(a), we use this parametrization and tune the coefficients $a_1,\dots,a_d$ during training (with some of them set to zero).

For the canonical dictionary $\D$, the algorithm prunes the DNN weights with the smallest absolute magnitudes.
It is worth highlighting that additional constraints can be imposed in Step 2(b) when pruning the dictionary, such as when using block-diagonal dictionaries as described later.
Also, some modifications can be added to the algorithm. For example, although we rewind to the original initialization, as per the original IMP, it has been suggested that rewinding may not be necessary when tuning the learning rate \cite{liu2018rethinking}.
Also, the computational burden of training for several times to convergence can be alleviated, e.g.\ by using a large learning rate to approximately fit the data at each round relatively quickly \cite{ding2019global,golub2018full,you2019drawing}.

\section{Examples}\label{sect:examples}

Before confirming the GLTH for two non-canonical choices of dictionaries, we discuss briefly how the choice of the dictionary $\D$ can be ``aligned'' with the DNN architecture.

\begin{figure}[b!]
    \centering
    \includegraphics[width=0.23\linewidth]{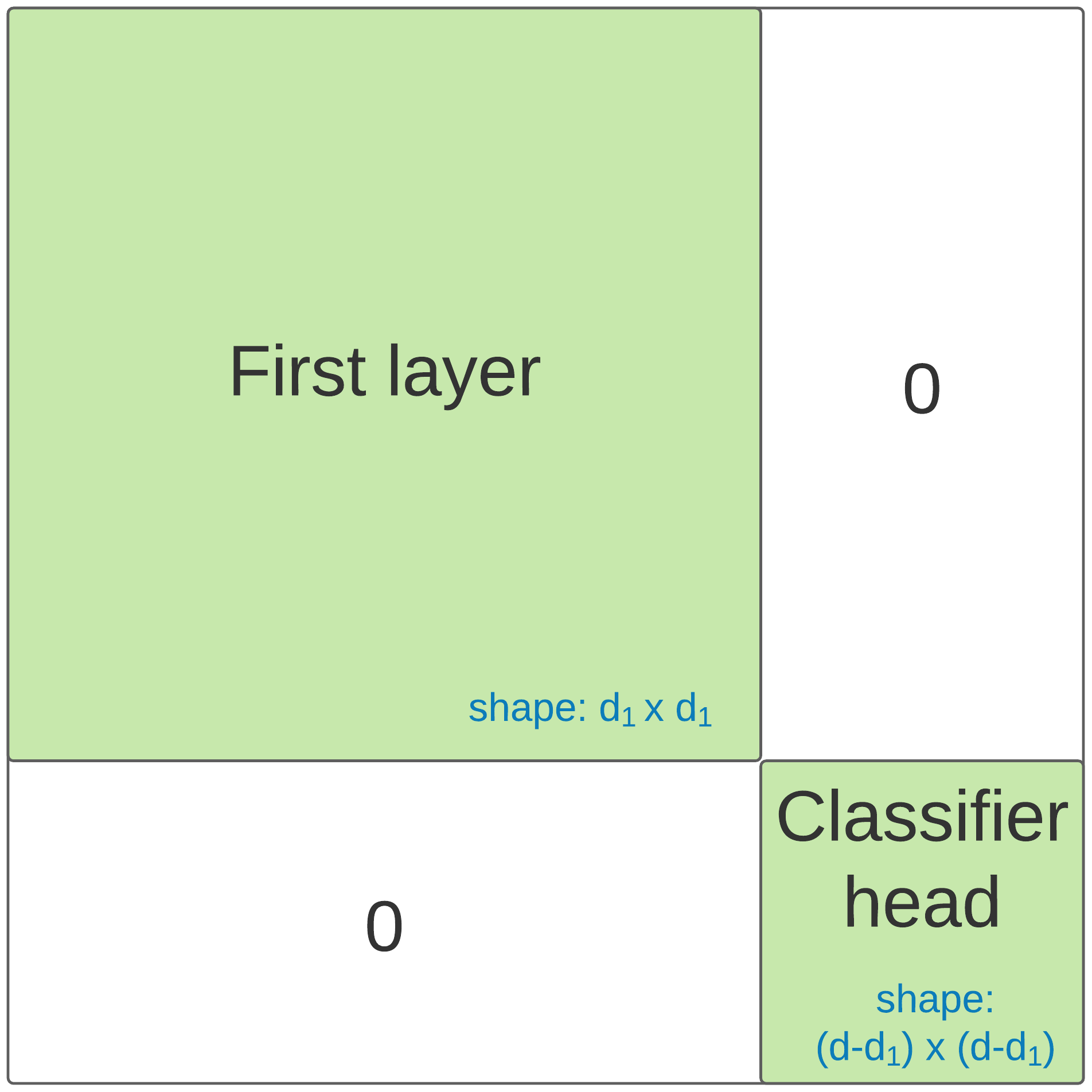}
    \hfill
    \includegraphics[width=0.23\linewidth]{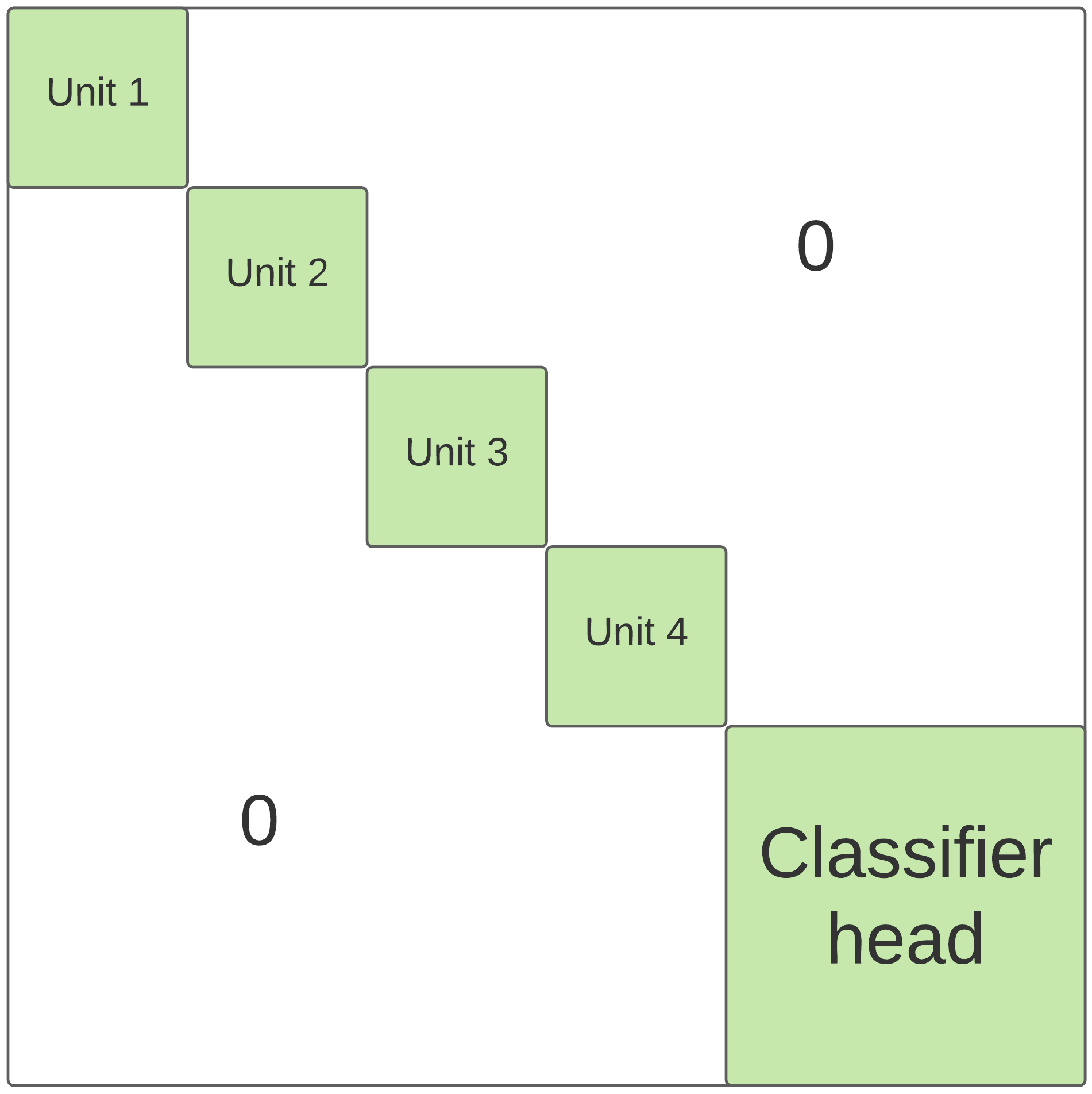}
    \hfill
    \includegraphics[width=0.227\linewidth]{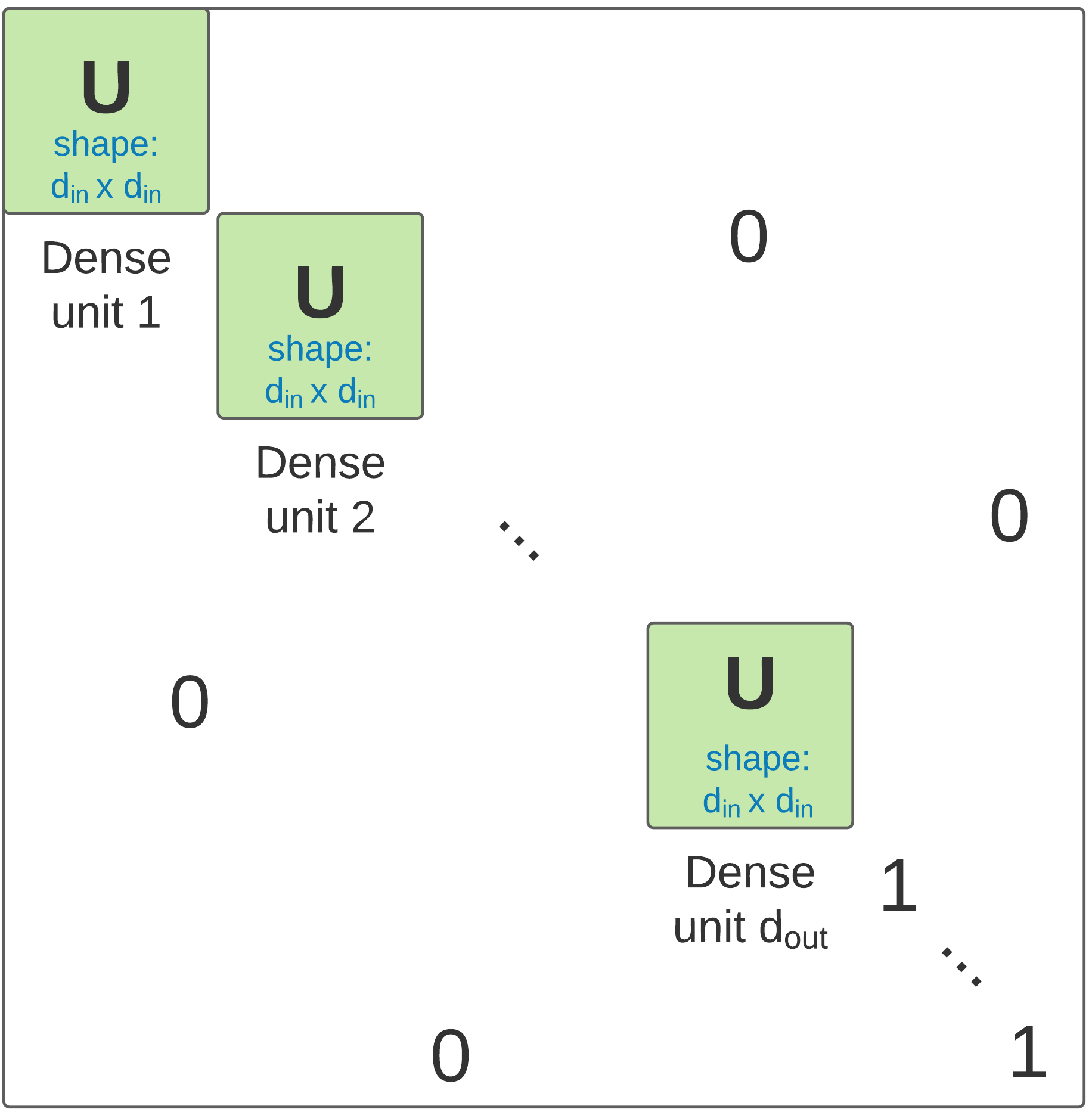}
    \hfill
    \includegraphics[width=0.23\linewidth]{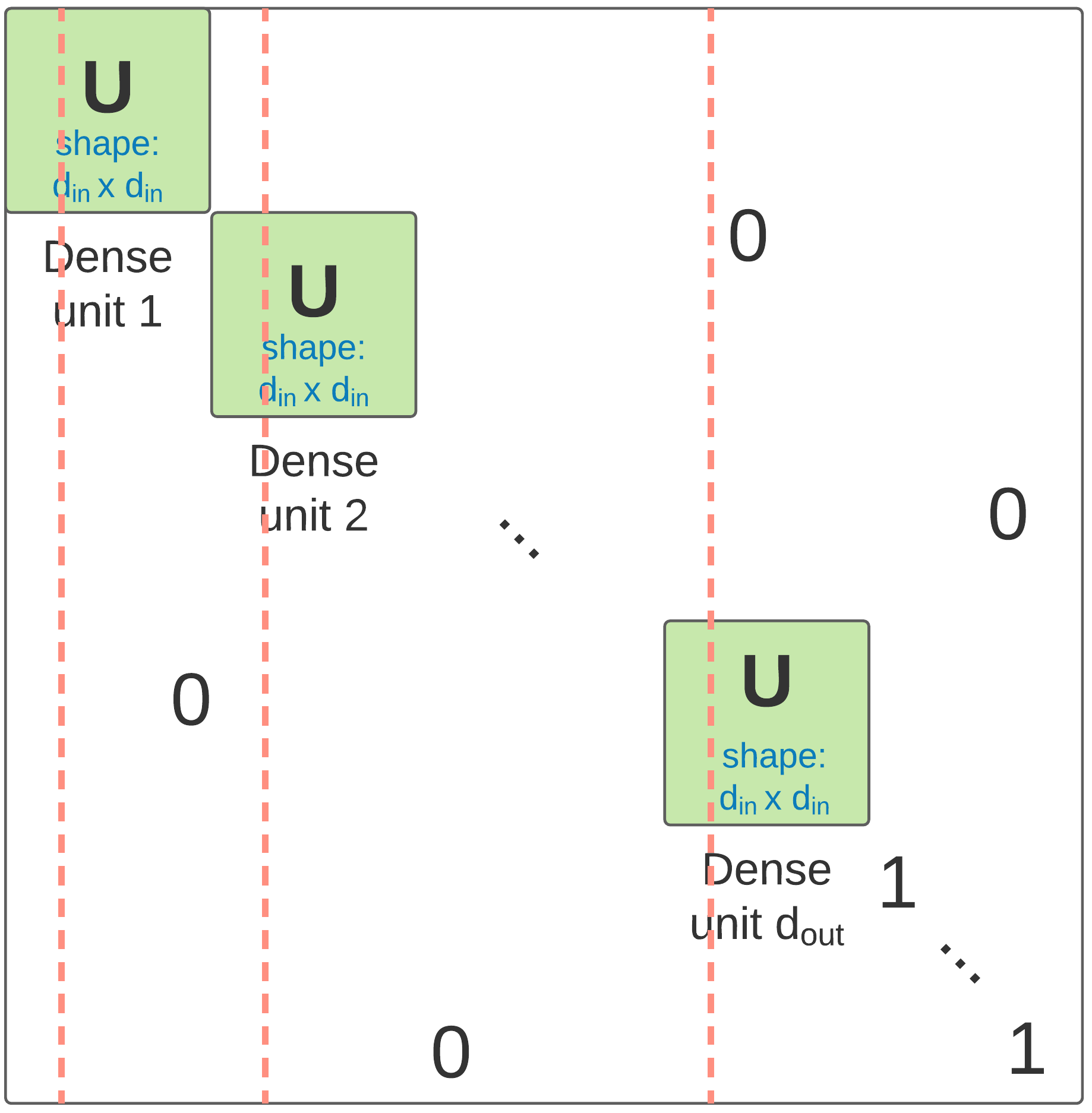}
    \caption{{\bf Left:} Block-diagonal dictionary matrix for the DNN architecture with one hidden layer followed by a classifier head. Re-parameterizing with respect to the corresponding dictionary allows to tune the first layer independently of the classifier head.
    {\bf Center-left:} Another dictionary for the same architecture that allows to tune separate dense units and the classifier head independently.
    {\bf Center-right:} A bottleneck dictionary for the dense layer with $\dout$ hidden units, in which an orthogonal matrix $\mathbf{U}$ is \emph{identical} across the units (see Section \ref{sect:examples}).
    {\bf Right:} When applying step 2.(b) with the bottleneck dictionary, we delete groups of columns (dotted lines in red) so that the \emph{same column} is removed from each of the diagonal blocks (that are shared).
    }
    \label{fig:two-layers}
\end{figure}

Keeping in mind that a dictionary $\D\subset \R^d$ can be represented as an orthogonal matrix in $\R^{d\times d}$ formed by stacking the column-vectors $\{v_1,\dots, v_d\}$ of the dictionary, alignment with the architecture is 
achieved using block-diagonal dictionaries as illustrated in Figure \ref{fig:two-layers}.
Re-parametrizing with such dictionaries allows to optimize different blocks of the DNN architecture separately. 

More precisely, suppose we have a DNN architecture with $d$ trainable parameters that comprises of one fully-connected layer and a classifier head as an example.
Let $w\in\R^d$ be the vector of all parameters, where the first $d_1$ weights correspond to the first layer.
Consider re-parametrizing with respect to the block diagonal dictionary illustrated in Figure \ref{fig:two-layers}({\sc left}).
Changing the new model parameters that correspond to the leftmost $d_1$ basis elements would only affect the first layer of the DNN, without impacting the classifier head. 
Figure \ref{fig:two-layers}({\sc center-left}) shows a different choice of block-diagonal dictionaries in which different \emph{hidden units} in the first layer can be optimized separately by dedicating a diagonal block to each individual unit.
Following this path to the extreme with $1\times 1 $ blocks leads to the canonical basis, in which every single model parameter can be tuned separately.

\begin{figure}[tb]
    \centering
    \includegraphics[width=0.45\linewidth]{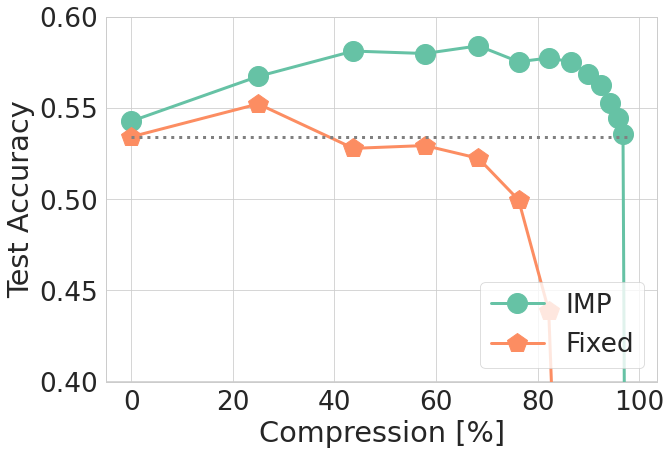}
    \includegraphics[width=0.45\linewidth]{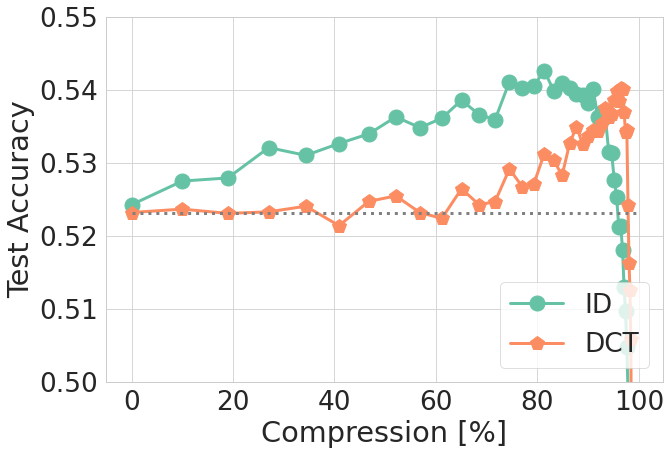}
    \caption{CIFAR10 test accuracy plotted against compression ratio when IMP is used for non-canonical dictionaries. As observed for the original Lottery Ticket Hypothesis, the test accuracy initially improves before it decreases. Also, there exists a critical threshold after which the accuracy drops rapidly. {\sc left:} A CNN (30K parameters) and a random dictionary. The green curve uses IMP to prune the basis while the orange curve corresponds to the setting of \citet{Yosinski2018} in which the subset of basis is fixed in advance. {\sc right:} An MLP with pruning applied to the 1\textsuperscript{st} layer (1M parameters) using bottleneck dictionaries with either identity blocks (ID) or DCT.}
    \label{fig:random_dict}
\end{figure}

\paragraph{Random Dictionary.}
First, we demonstrate the GLTH on random dictionaries, in which $\D$ is a random rotation of the canonical basis. Unlike in the canonical basis, where removing a single component of $\D$ corresponds to eliminating a single parameter, removing a single component in a random dictionary affects \emph{all} the parameters in the neural network. We take a convolutional neural network with three hidden layers (16 filters $\mapsto$ 8 filters  $\mapsto$ 4  filters), where all filters are $3\times 3$, followed by a classifier head and we train it on CIFAR10. The total number of parameters is 28,950. 
We apply the IMP algorithm described earlier and for every iteration report the compression rate (i.e.\ ratio of removed elements in the dictionary over the original size) and the test accuracy. Figure~\ref{fig:random_dict}({\sc left}) reports the results.
As shown in the figure, the IMP algorithm with rewinding successfully finds a sparse solution whose accuracy is comparable to vanilla training, verifying the GLTH for this choice of the dictionary.

While sparsity with respect to such a particular choice of $\D$ does not have a useful  semantic meaning, it serves as a proof of concept to the GLTH.
It also nicely connects the literature on the lottery ticket hypothesis to the results of \citet{Yosinski2018}.
The authors train DNN models in \emph{randomly selected} $s$-dimensional subspace of the full parameter space $\R^d$.
They demonstrate that for various DNN architectures and datasets one can recover up to 90\% of the test accuracy obtained with vanilla training using surprisingly small values of $s$.
In contrast, we demonstrate that starting from \emph{any random set} of $d$ orthonormal vectors in $\R^d$ one can find a subset of $s$ vectors, such that their linear span contains a solution as good as the one obtained with vanilla training.
The orange curve in Figure~\ref{fig:random_dict}({\sc left}) corresponds to the setting of \citet{Yosinski2018} and demonstrates the performance with respect to the randomly chosen subspace. It also corresponds to the ``\emph{randomly sampled sparse networks}'' in Figure~1 of \citet{frankle2018lottery}.

\paragraph{Bottleneck Dictionary.}
Next we introduce \emph{bottleneck dictionaries}.
We verify the GLTH for these dictionaries and show that when combined with a slight modification of the IMP algorithm, they lead to bottleneck layers.
A bottleneck layer in this context is a fully-connected layer in which the weight matrix $\mathbf{W}\in\R^{\din \times \dout}$ can be  factored into a product of two low-rank matrices.  Here, $\din$ is the number of input units while $\dout$ is the number of hidden units in the fully-connected layer. 

A bottleneck dictionary re-parametrizes the weights in $\mathbf{W}$, while keeping all the other parameters of the DNN (including the bias terms of the  dense layer) intact.
Slightly overloading our earlier notation, in the rest of this paragraph we will write $w\in \R^d$ to denote the flattened vector of $d=\din \times \dout$ weight parameters of the matrix $\mathbf{W}$.
The orthogonal matrix corresponding to the dictionary will have one $d\times d$ block (for the weight matrix of the dense layer) and many $1\times 1$ blocks (for each of the remaining trainable DNN parameters, including the bias terms of the dense layer).

Now it is left to describe the diagonal block of shape $d\times d$.
As shown in Figure \ref{fig:two-layers} ({\sc center-right}), we will set this block to comprise itself of  $\dout$ smaller diagonal blocks, one for each of the $\dout$ hidden units in that fully-connected layer.
In addition, they are all \emph{shared}: each of the smaller blocks is equal to the \emph{same} orthogonal matrix $\mathbf{U} \in \R^{\din \times \din}$ with $\mathbf{U} \mathbf{U}^\top = I$.
This re-parametrization means that 
every hidden unit (column of matrix $\mathbf{W}$) is expressed as a linear combination of the columns of $\mathbf{U}$. 
Different units share the same basis $\mathbf{U}$, but have different coefficients:
\[
\mathbf{W} = \mathbf{U} \mathbf{C},\quad \mathbf{U}\in\R^{\din \times \din}, \mathbf{C} \in \R^{\din \times \dout}.
\]

We have just defined the dictionary $\DBN$ that we call a \emph{bottleneck dictionary}. Clearly, this  dictionary is complete, i.e.\:the columns in the matrix depicted in Figure \ref{fig:two-layers} ({\sc center-right}) form a basis in the space of all trainable parameters of the neural network.

Next we discuss an application of the IMP algorithm to the bottleneck dictionary.
To make our subsequent discussion precise, we assume the following ordering of the parameters in the flattened vector $w\in\R^d$: the first $\din$ components of $w$ correspond to the first hidden unit, the second $\din$ components correspond to the second hidden unit, and so on.
Finally, to ensure that we have the desired low-rank decomposition of the weight matrix $\mathbf{W}\in\R^{\din \times \dout}$, we modify the IMP algorithm so that the diagonal blocks $\mathbf{U}$ (see Figure \ref{fig:two-layers} ({\sc center-right})) \emph{are shared}. In other words, if we remove a column from one of the $\mathbf{U}$ matrices on step 2.(b) of the IMP algorithm, we remove the same column from \emph{all other} $\mathbf{U}$ blocks (Figure \ref{fig:two-layers}({\sc right})).

More precisely, writing $\D = \{v_1,\ldots,v_d\}$ where $d=\din\times\dout$, we modify Step 2.b to be:

\begin{enumerate}
    \item[2.(b)]
    ({\bf Sparsify}) Remove one (or more) elements from $\D_t$ resulting in $\D_{t+1}\subsetneq\D_t$, where: $\D_{t+1} = \D_t \setminus \tilde{\D}(\ell)$ and
    $\tilde{\D}(\ell):=\{v_{k \cdot \din + \ell}:\; k=0,\dots,\dout - 1\}$
    for $\ell \in \{1,\dots, \din\}$.
    We choose $\ell$ to minimize, again, the Euclidean distance between $w_t$ and its projection into the new subspace $P_{\textbf{Span}(\D_{t+1})}w_t$.
\end{enumerate}
Upon termination, reshaping the final parameters $w\in\R^d$ into $\mathbf{W}\in\R^{\din \times \dout}$ will result in a matrix that can be factorized into a product of two low-rank matrices: \begin{equation}
\label{eq:bottleneck}
\mathbf{W} = \mathbf{U}' \mathbf{C}',\quad \mathbf{U}'\in\R^{\din \times m}, \mathbf{C}' \in \R^{m \times \dout},
\end{equation}
where $\mathbf{U}'$ contains the $m$ remaining columns from $\mathbf{U}$ after pruning, while $\mathbf{C}'$ are trained coefficients. 

Next we discuss two different choices of the matrix $\mathbf{U}$.
The {\bf first option} is the identity block $\mathbf{U}=I\in\R^{\din \times \din}$.
In this case step 2.(b) of the IMP algorithm above forces the dense layer to entirely \emph{ignore} some of the input units.
This can be used either to ignore most of the pixels in the input image (if the dense layer is the  first layer in the DNN) or to prune the hidden units from the {previous dense layer}.
The {\bf second option} is a $\mathbf{U}$ matrix where the columns form the DCT basis in the space of $32\times 32$ images.
This option makes sense when pruning the \emph{first} dense layer of the DNN that receives the (flattened) images as inputs.
In case the input images have more than one channel (e.g.\ RGB for CIFAR10) we use one DCT basis per each of the input channels.

\begin{figure}[t]
    \centering
    \includegraphics[width=0.2\linewidth]{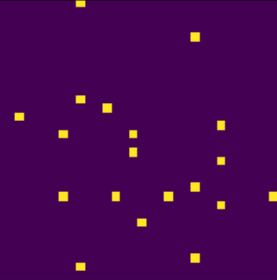}
    \hspace{.1cm}
    \includegraphics[width=0.2\linewidth]{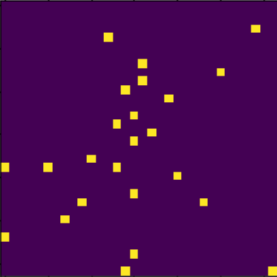}
    \hspace{.1cm}
    \includegraphics[width=0.2\linewidth]{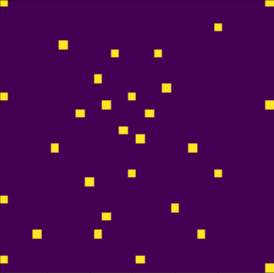}
    \hspace{.1cm}
    \includegraphics[width=0.2\linewidth]{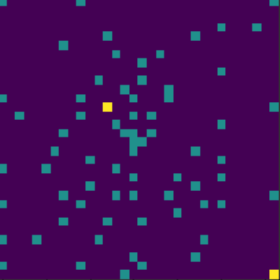}\\
    \vspace{.2cm}
    \caption{Running the IMP algorithm with the identity bottleneck dictionary in the first layer prunes entire input coordinates (i.e. channel-pixels in images). 
    Left-to-right: remaining channel pixels in the R, G, B channels and sum of all 3. Every picture is $32\times 32$. 
    }
    \label{fig:bottleneck_identity}
\end{figure}

\begin{figure}[tbp]
    \centering
    \includegraphics[width=1.\linewidth]{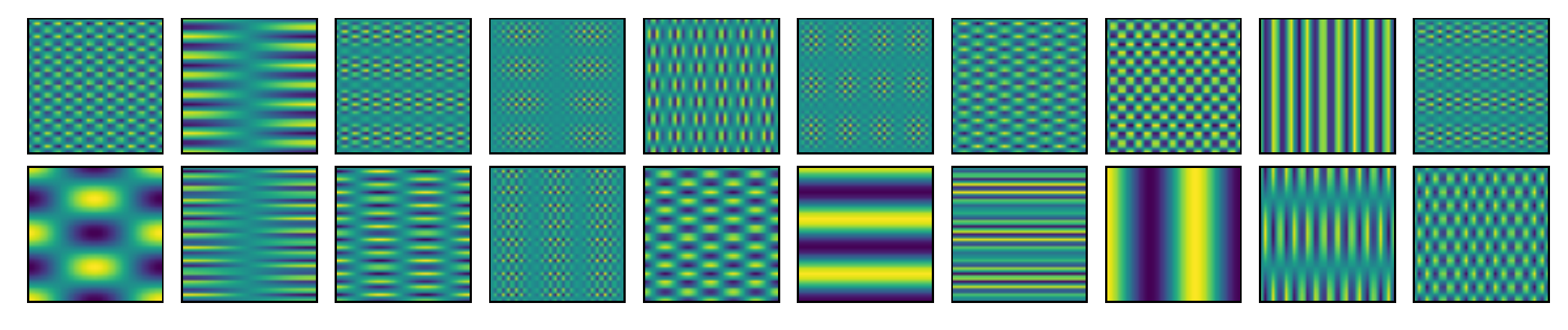}
    \caption{Randomly chosen elements from the DCT basis in the $32\times 32$ image space, which contains  1024 elements.
    We use one DCT basis per each of the R, G, and B input channels.}
    \label{fig:dct}
\end{figure}

We report experiments with both options in Figure \ref{fig:random_dict} ({\sc right}), where we use a network with two hidden dense layers (300 units $\mapsto$ 100 units $\mapsto$ classifier head) and train it on CIFAR10.
We prune \emph{only the first layer} that has $32\times 32 \times 3 \times 300 \approx 1\text{M}$ parameters and report its compression rate, where ``100\%'' corresponds to pruning entire first layer.
The plot confirms the GLTH for the bottleneck dictionaries.
Interestingly, we can preserve the original test accuracy of $\sim 52$\% for the vanilla training with a model that ignores all but $\sim 150$ input channel-pixel values (which is $\sim 5$\% out of $32\times 32 \times 3 = 3072$ overall pixel values).
The coordinates (channel-pixels) used by the pruned model are shown in Figure \ref{fig:bottleneck_identity}.
Finally, using the DCT basis in $\mathbf{U}$ provides an even higher compression rate ($\sim 98$\% of the parameters pruned) for the same level of accuracy. 
Examples of the DCT basis are shown in Figure \ref{fig:dct}. 

Finally, we briefly discuss the actual computational benefits when pruning with the bottleneck dictionaries. 
If $m \ll \min(\din, \dout)$ in \eqref{eq:bottleneck} this low-rank factorization leads to significant increase in the inference speed. 
Additionally, when using identity $\mathbf{U}$ we can also remove hidden neurons from the \emph{previous} dense layer, which further increases the inference speed.

The low-rank decomposition of the weight matrix \eqref{eq:bottleneck} is discussed in detail in \citet{denil2013predicting}.
The authors focus on reducing the amount of trainable parameters in DNNs.
Given a fixed number $m$, they propose various ways of choosing the $m$ columns in $\mathbf{U}'\in\R^{\din \times m}$ so that training parameters in $\mathbf{C}'$ results in good generalization.
In this work we choose the orthonormal basis in $\mathbf{U}\in\R^{\din \times \din}$ and iteratively remove the columns in $\mathbf{U}$, following the IMP algorithm.



\section{Acknowledgements}
We would like to thank Olivier Bousquet, Hartmut Maennel, Robert Baldock, Lucas Beyer, Sylvain Gelly, Neil Houlsby, and the  Google Brain team at large for their support.

\bibliographystyle{abbrvnat}  
\bibliography{glt2021bibliography}
\end{document}